\def\year{2019}\relax
\documentclass[letterpaper]{article} 
\usepackage{aaai19}  
\usepackage{times}  
\usepackage{helvet}  
\usepackage{courier}  
\usepackage{url}  
\usepackage{graphicx}  
\usepackage{amssymb}
\usepackage{xspace}
\usepackage{makecell}
\usepackage{amsmath}
\usepackage{url}
\usepackage{hyperref}
\usepackage{multirow}
\usepackage{graphicx}  
\usepackage{algorithm,algorithmicx,algpseudocode}
\usepackage{booktabs}
\usepackage{enumitem}
\usepackage{bbm}

\makeatletter
\def\BState{\State\hskip-\ALG@thistlm}
\makeatother

\newcommand{\expect}[1]{\ensuremath{\underset{#1}{\mathbb{E}}\xspace}}

\begin{document}
%
\title{Approximate Distribution Matching for Sequence-to-Sequence Learning}
\author{Wenhu Chen\thanks{The first two authors finished the paper in Microsoft as interns}$^\dag$, Guanlin Li,$^\S$, Shujie Liu,$^\ddag$ Zhirui Zhang,$^\P$ Mu Li,$^\ddag$, Ming Zhou$^\ddag$\\ 
$^\dag$University of California, Santa Barbara\\
$^\S$Harbin Institute of Technology\\
$^\P$University of Science and Technology of China\\
$^\ddag$Microsoft Research Asia\\
\texttt{wenhuchen@cs.ucsb.edu}, \texttt{epsilonlee.green@gmail.com} \\
\texttt{\{shujliu, v-zhirzh, muli, mingzhou\}@microsoft.com}
}
\maketitle
\begin{abstract}
Sequence-to-Sequence models were introduced to tackle many real-life problems like machine translation, summarization, image captioning, etc. The standard optimization algorithms are mainly based on example-to-example matching like maximum likelihood estimation, which is known to suffer from data sparsity problem. Here we present an alternate view to explain sequence-to-sequence learning as a distribution matching problem, where each source or target example is viewed to represent a local latent distribution in the source or target domain. Then, we interpret sequence-to-sequence learning as learning a transductive model to transform the source local latent distributions to match their corresponding target distributions. In our framework, we approximate both the source and target latent distributions with recurrent neural networks (augmenter). During training, the parallel augmenters learn to better approximate the local latent distributions, while the sequence prediction model learns to minimize the KL-divergence of the transformed source distributions and the approximated target distributions. This algorithm can alleviate the data sparsity issues in sequence learning by locally augmenting more unseen data pairs and increasing the model's robustness. Experiments conducted on machine translation and image captioning consistently demonstrate the superiority of our proposed algorithm over the other competing algorithms. 
\end{abstract}

\section{Introduction}
Deep learning has achieved great success in recent years, especially in sequence-to-sequence applications like machine translation~\cite{bahdanau2014neural,cho2014learning}, image captioning~\cite{xu2015show,DBLP:conf/cvpr/RennieMMRG17}, abstractive summarization~\cite{DBLP:conf/emnlp/RushCW15,paulus2017deep} and speech recognition~\cite{wu2016stimulated,lu2015study}, etc. The most common approaches are based on neural networks which employ very large parameter set to learn a transductive function between the input space and the target space. 

The key problem faced by neural sequence-to-sequence model is how to learn a robust transductive function in such a high-dimensional space with rather sparse human-annotated data pairs. For example, machine translation takes the input sequence $x$ which lies in the space of $|\mathbb{V}|^N$ to output sequence in another $|\mathbb{\hat{V}}|^{\hat{N}}$ space, where $\mathbb{V},\mathbb{\hat{V}}$ denote the vocabulary sizes and $N,\hat{N}$ denote the sequence lengths. In the large-scale problem, the input and output space become so large that any amount of annotated dataset appears to be sparse. Such data sparsity problem poses great challenges for the model to understand both the input and output diversity. It's worth noting that our claimed data sparsity problem is specific to sequence-to-sequence scenario, which is slightly different from \textit{curse of dimensionality}\footnote{\small The curse of dimensionality problem happens when the feature dimension is too high for the limited data to fit while our claimed data sparsity happens when the data space is too large for the limited data to cover.}. In general, our method runs parallel with the methods which prevent model overfitting like l1/2 regularization, dropout~\cite{srivastava2014dropout}, and etc.

\begin{figure}[htb]
\begin{center}
\includegraphics[width=1.0\linewidth]{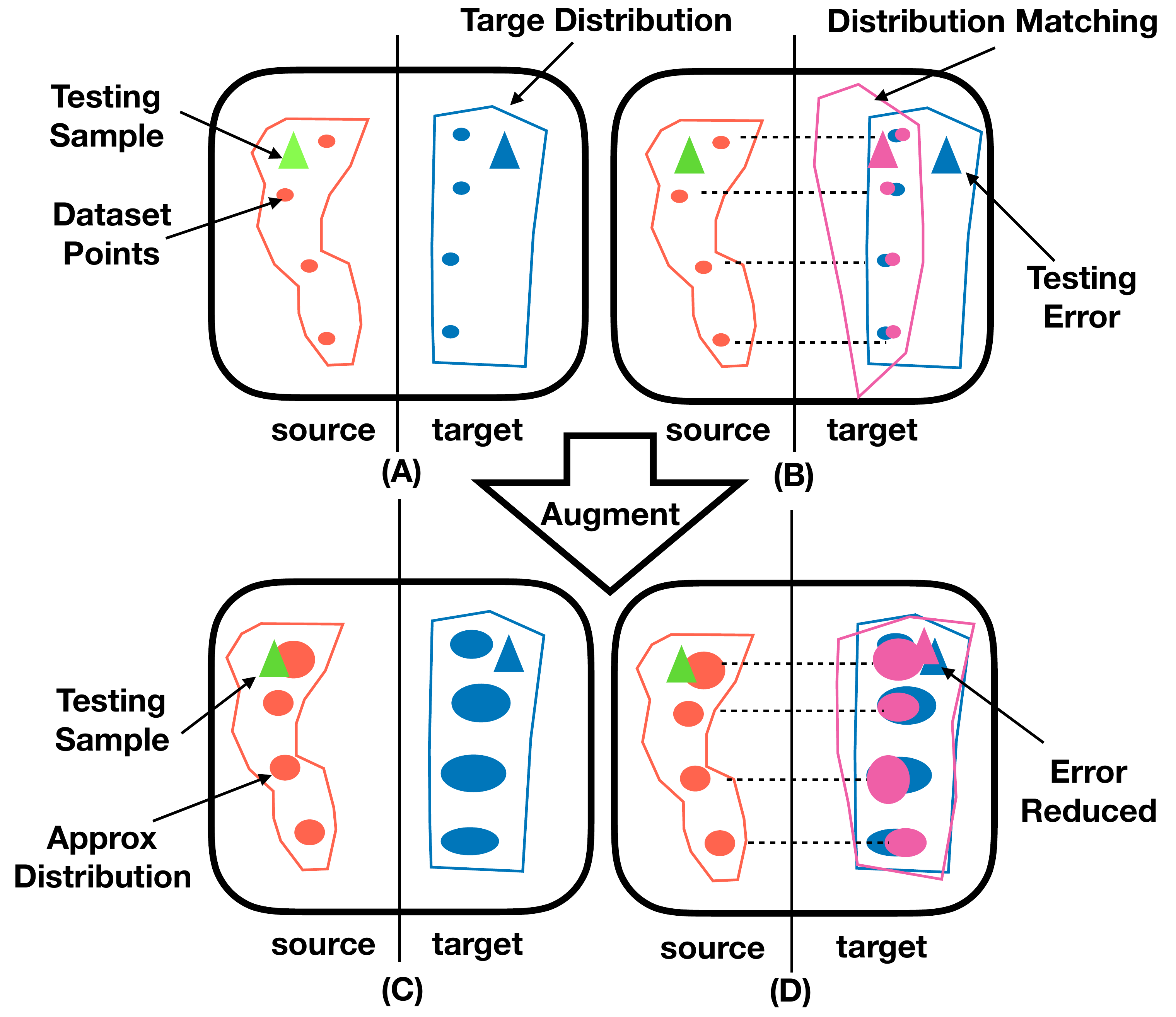}
\end{center}
   \caption{Illustration of our method: (A) There are two true distributions in the source and target space, red means the source sequence and blue means the target sequence. Here we assume there are four training sample pairs (dots) and one testing sample (triangles) in the dataset. (B) We use the sequence-to-sequence model to maximize the likelihood of target data points given source data points. Though the alignments of four dots are perfect, the overall space is misaligned causing errors in testing samples. (C) We use the augmenters to expand dots as ellipses for both sides. (D) The sequence-to-sequence model can align the overall source and target space better, thus minimizing the error for testing samples.}
\label{fig:TATA}
\end{figure}

In order to resolve the specific data sparsity problem in sequence-to-sequence learning, different data augmentation approaches~\cite{DBLP:conf/acl/SennrichHB16,ma2017softmax,norouzi2016reward,he2016dual,chen2018generative} have been proposed. These methods are mainly focused on ``augmenting'' pseudo-parallel data to fully explore the data space, their main weaknesses can be mainly summarized into the following aspects: 1) Back-Translation~\cite{DBLP:conf/acl/SennrichHB16} are specific to certain task like NMT; 2) Reward-Augmented Training~\cite{ma2017softmax,norouzi2016reward} fail to consider source side diversity; 3) Dual Learning~\cite{he2016dual} requires duality property and additional resources.

In this paper, we are devoted to design a general-purpose sequence-to-sequence learning algorithm to alleviate data sparsity problem without relying on any external resources. We first assume every example in the dataset actually represents an unknown latent distribution, which we need to approximate. In the language domain, the latent distribution could be viewed as a set of paraphrases, while in the image domain, the latent distribution could be thought of as a set of similar pictures. The current prevalent heuristics for approximating the latent distribution~\cite{norouzi2016reward} are mainly based on token-level replacement, which are known to suffer from the following problems: 1) Inconsistency: RAML~\cite{norouzi2016reward} does not retain the fidelity to original data pairs and breaks the pairwise correspondence\footnote{\small RAML could turn an English sentence from ``a girl is going to \texttt{school}" into ``a girl is going to \texttt{kitchen}" while a German translation from ``ein(a) \texttt{Mädchen(girl)} gehe(goes) ...'' to ``ein(a) \texttt{Junge(boy)} gehe(goes) ...''.}. 2) Broken Structure: paraphrasing potentially breaks the structure of the sequence and causes unnecessary errors\footnote{Heuristic replacement could turn a source sentence from ``a girl \texttt{is going to} school''  into ``a girl \texttt{plans to} school''.}. 3) Discreteness: these methods are merely used for a sequence with discrete tokens, not suitable for a sequence with continuous vector scenario. 

In order to defeat these issues to augment fluent and well-corresponded source-target pairs, we design our system to meet the following three criterion: 1) \textit{generability}: we employ the generative model (augmenter) to generate new sequences rather than using hard replacements, which can avoid broken structure and be applicable to continuous variable scenarios; 2) \textit{fidelity}: we restrict the augmented pair to follow their original prototype by maximizing their likelihood computed by the sequence model; 3) \textit{diversity}: we encourage the augmenters to output more unseen samples to cover the large data space. These designs can enable the augmenters to better approximate the latent distributions, which then enhances the robustness of sequence-to-sequence learning. A pedagogical illustration is shown in~\autoref{fig:TATA}, where we learn the latent distribution and then employ the sequence model to align them. The testing error can be reduced by fully exploring the data space.

In conclusion, the major contributions of our paper are described as follows: 
\begin{itemize}[noitemsep,topsep=0pt]
\item we are the first to view sequence-to-sequence learning as a distribution matching problem.
\item we have successfully applied our algorithm into two large-scale real-life tasks and design corresponding architectures for them.
\item we have empirically demonstrated that our method can remarkably outperform the existing algorithms like MLE, RL, and RAML.
\end{itemize}

\section{Related Literature}
\subsection{Neural Sequence Model}
A major recent development in machine learning community is the adoption of neural networks. Neural network models promise better sharing of statistical evidence between similar words and inclusion of rich context. Since~\cite{bahdanau2014neural,cho2014learning} proposed the sequence-to-sequence model, it has been widely adopted in the industries and academia. Later on, many follow-up works on machine translation like~\cite{chen2016guided,wu2016google} and visual captioning~\cite{xu2015show,DBLP:conf/cvpr/RennieMMRG17,chen2016bootstrap} have been proposed to achieve  state-of-the-art performance. 


\subsection{Reinforcement Learning}
Exposure bias and train-test loss discrepancy are two major issues in the training of sequence prediction models in neural machine translation or image captioning. Many research works ~\cite{bahdanau2014neural,ranzato2015sequence,xu2015show,DBLP:conf/cvpr/RennieMMRG17} have attempted to tackle these issues by exposing the model to its own distribution and directly maximizing task-level rewards. These methods are reported to achieve significant improvements in many applications like machine translation, image captioning and summarization, etc. These works are able to encourage the sequence model to exploit the target space better by driving it with a human-crafted reward signal, our method can also encourage the sequence model to exploit the source and target space with a sophisticated model-based reward signal.

\subsection{Reward Augmented Training}
One successful approach for data augmentation in neural machine translation system is RAML~\cite{norouzi2016reward}, which proposes a novel payoff distribution to augment training samples based on task-level reward (BLEU, Edit Distance, etc). In order to sample from this intractable distribution, they further stratify the sampling process as first sampling an edit distance, then performing random substitution/deletion operations. In order to combat the unnecessary noises introduced by the random replacement strategy, our method considers semantic and syntactic context to perform paraphrase generation.  

\section{Preliminary}
Here we first introduce the sequence-to-sequence model proposed in~\cite{cho2014learning,bahdanau2014neural}, which applies two recurrent neural networks~\cite{mikolov2010recurrent} to separately understand input sequence and generate output sequence. This framework has been widely applied in various sequence generation tasks due to its simplicity and end-to-end nature, which successfully avoids expensive human-crafted features. The sequence model receives the feedback and form a distribution $z$ over the output space according to chain rule as follows:
\begin{align}
\begin{split}
&x_t \sim z(r_t, c_t)\\
&r_t = g(r_{t-1}, x_{t-1}, c_t)\\
&c_t = q(r_{t-1}, (h_1, \cdots, h_T))
\end{split}
\end{align}
where $r_t$ are the recurrent units, $q$ is a global attention function to compute the attention weights $c_t$ over the input information $h$. For generality, the sequence element $x_t$ could be a discrete integer or a real-value vector depending on the distribution $z$. In language related task, $x_t$ lies in the discrete space $x_t \in \{1, \cdots, V\}$, where the most frequently used is the Multinomial distribution:
\begin{align}
\begin{split}
    &x_t \sim Multinomial(P = f_{p}(r_t, c_t)) \\
    &where \qquad \sum_i P_i = 1
\end{split}
\end{align}
where $p \in \mathbb{R}^V$ is the output of function $f_{p}$.

In contrast, in visual captioning, $x_t$ can be seen as the representation of image lying in the continuous d-dimensional space $x_t \in \mathbb{R}^d$, where the most popular option is multivariate Gaussian distribution:
\begin{align}
\begin{split}
    &x_t \sim \mathbb{N}(\mu=f_{\mu}(r_t, c_t) , \sigma^2=f_{\sigma}(r_t, c_t)) \\
\end{split}
\end{align}
where $\mu \in \mathbb{R}^d, \sigma \in \mathbb{R}^d$ are the Gaussian mean and deviation obtained from functions $f_{\mu}, f_{\sigma}$. We will cover these two cases in the following sections.
\begin{figure}[htb]
\begin{center}
\includegraphics[width=1.0\linewidth]{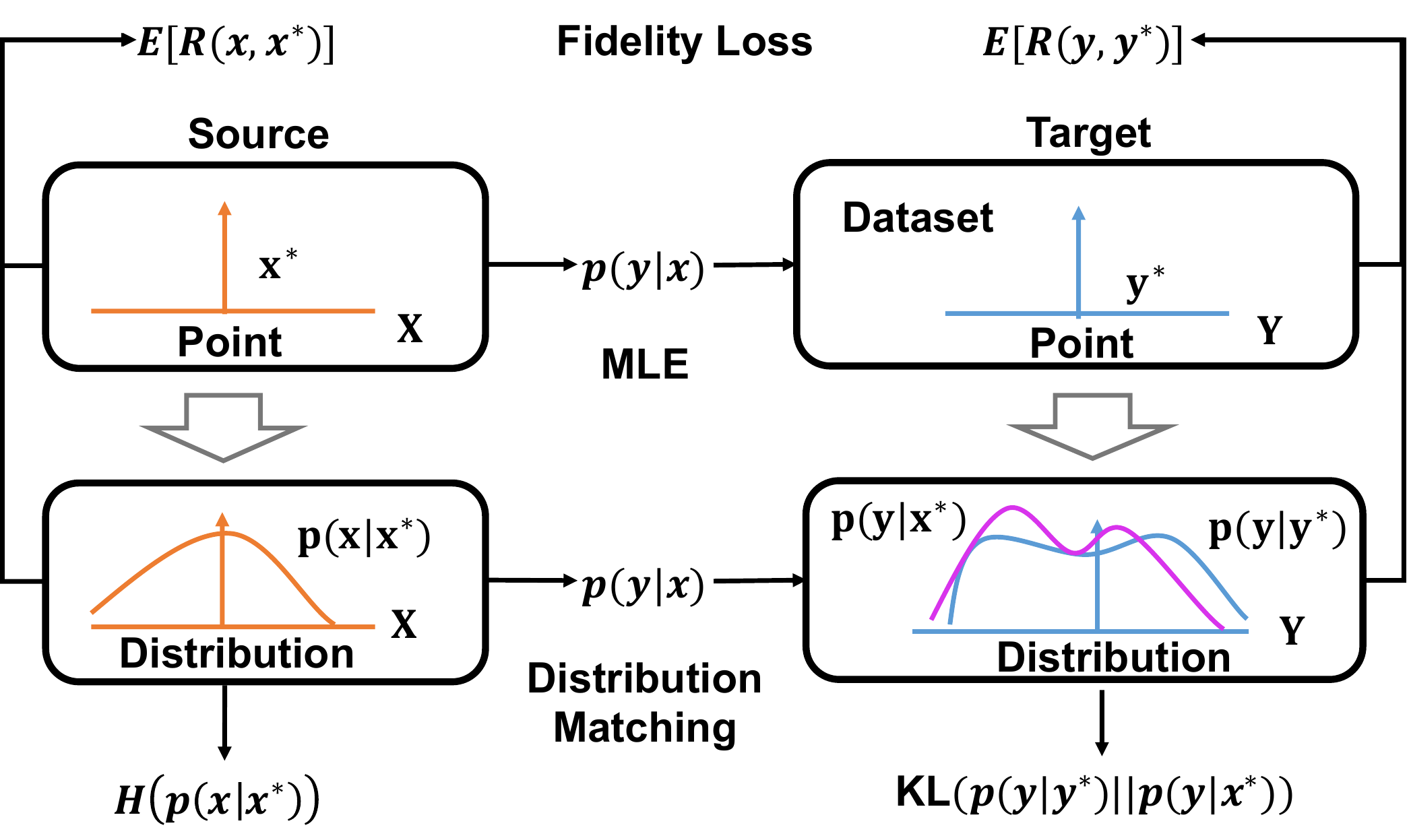}
\end{center}
   \caption{Illustration of our objective function: both source and target augmenters are expanding their own data samples into distributions. The sequence model transforms the source distribution into the target domain, which is matched against augmented target distribution by minimizing the KL-divergence. The additional entropy regularization is designed to increase output diversity. }
\label{fig:TATA-dist}
\end{figure}

\section{Model}
\subsection{Overview}
Here we demonstrate our philosophy using a pedagogical illustration in~\autoref{fig:TATA}, the example demonstrates how our distribution-matching framework works in combating data sparsity problem to improve model's ability to understand the diversity in both sides.
Our framework first introduces the parallel augmenter, which views the source-target pairs $(x^*, y^*)$ from the dataset as a prototype and aims at augmenting them simultaneously to output synthetic pairs $(x, y)$. Specifically, we parameterize the source side and target side augmenters as $p_{\theta}(x|x^*)$ and $p_{\gamma}(y|y^*)$, which are also implemented with recurrent neural networks. Then we elaborate the above mentioned constraints (see introduction) into two objective functions:
\begin{itemize}
\item Matching loss: the transformed source distribution $p_{\theta, \beta}(y|x^*)$ should match its corresponding local latent distribution $p_{\gamma}(y|y^*)$ in the target domain.
\begin{align}
\small
J_{match} = &\expect{(x^*,y^*) \sim D} [-KL(p_{\gamma}(y|y^*)||p_{\theta, \beta}(y|x^*))]
\end{align}
where we use $p_{\theta, \beta}(y|x^*)$ to denote the marginal likelihood $\sum_{x} p_{\theta}(x|x^*)p_{\beta}(y|x)$. However, we found that such KL-divergence can degenerate into Maximum Likelihood Estimation by setting $p_{\gamma}(y|y^*), p_{\theta}(x|x^*)$ to Kronecker-delta function $\delta(x, x^*)$. Such scenario will violate the diversity constraint, therefore, we leverage an entropy regularization term in the source side to avert that. The matching loss can hence be expressed as follows:
\begin{align}
\begin{split}
\small
J_{match} = &\expect{(x^*,y^*) \sim D} [-KL(p_{\gamma}(y|y^*)||p_{\theta, \beta}(y|x^*))] + \\ &\expect{(x^*,y^*) \sim D} [H(p_{\theta}(x|x^*))]
\end{split}
\end{align}
\item Fidelity Loss: the randomly drawn samples should remain fidelity to its own ground truth.
\begin{align}
\begin{split}
\small
J_{fidelity} = &\expect{(x^*,y^*) \sim D} [\expect{x \sim p_{\theta}(x|x^*)} [-R(x, x^*)]] + \\
&\expect{(x^*,y^*) \sim D} [\expect{y \sim p_{\gamma}(y|y^*)} [-R(y, y^*)]]
\end{split}
\end{align}
where $R(x, x^*)$ denotes the similarity score (e.g. BLEU, METEROR in discrete case, or other distance measure in continuous case).
\end{itemize}
With the above two loss function, we propose to sum them as the combined loss function as follows:
\begin{align}
J=J_{match} + J_{fidelity}
\end{align}
Here we draw a pedagogical illustration of our proposed objective function in~\autoref{fig:TATA-dist}. During optimization, we will optimize the joint loss function directly with stochastic gradient descent. 

\subsection{Optimization}
Formally, we first write gradient of matching loss with respect to two augmenters and the sequence model as follows:
\begin{align}
\small
\begin{split}
&-\frac{\partial J_{match}}{\partial \gamma} = \expect{y \sim p_{\gamma}} [\log \frac{p_{\gamma}(y|y^*)}{p_{\theta, \beta}(y|x^*)} \frac{\partial \log p_{\gamma}(y|y^*)}{\partial \gamma}]\\
&-\frac{\partial J_{match}}{\partial \theta} = \expect{x \sim p_{\theta}} [\expect{y \sim p_{\gamma}}[\frac{p_{\beta}(y|x)}{p_{\theta, \beta}(y|x^*)}] \frac{\partial \log p_{\theta}(x|x^*)}{\partial \theta}]\\
& \qquad \qquad + \expect{x \sim p_{\theta}} [\log p_{\theta}(x|x^*) \frac{\partial \log p_{\theta}(x|x^*)}{\partial \theta}]\\
&-\frac{\partial J_{match}}{\partial \beta} = \expect{y \sim p_{\gamma}} [\expect{x \sim p_{\theta}} [\frac{p_{\beta}(y|x)}{p_{\theta, \beta}(y|x^*)} \frac{\partial \log p_{\beta}(y|x)}{\partial \beta}]]
\end{split}
\end{align}
Here we adopt Monte-Carlo algorithm to approximate the gradients as follows: 1) sample N source sequence samples $\{x^i\}_{i=1}^N$ and N target sequence samples $\{y^i\}_{i=1}^N$ from augmenters. 2) estimate $p(y^i|x^*)$ with $\frac{1}{N} \sum_{j=1}^N p(y^i|x^j)$. 3) use the sampled source and target sequences to estimate the gradients as follows:
\begin{align*}
\small
\begin{split}
-\frac{\partial J_{match}}{\partial \gamma} &\approx \frac{1}{N} \sum_{i=1}^N \log \frac{p_{\gamma}(y^i|y^*)}{p(y^i|x^*)} \frac{\partial \log p_{\gamma}}{\partial \gamma}\\
-\frac{\partial J_{match}}{\partial \theta} &\approx \frac{1}{N} \sum_{j=1}^N [\frac{1}{N} \sum_{i}^{N} \frac{p_{\beta}(y^i|x^j)}{p(y^i|x^*)} + \log p(x^j|x^*)] \frac{\partial \log p_{\theta}}{\partial \theta}\\
-\frac{\partial J_{match}}{\partial \beta} &\approx \frac{1}{N} \sum_{i}^N[\sum_{j}^N \frac{p_{\beta}(y^i|x^j)}{p(y^i|x^*)} \frac{\partial \log p_{\beta}}{\partial \beta}] 
\end{split}
\end{align*}
Then we write gradient of fidelity loss with respect to two augmenters as follows:
\begin{align}
\small
\begin{split}
-\frac{\partial J_{fidelity}}{\partial \gamma} &\approx \frac{1}{N} \sum_{i=1}^N R(y^i,y^*) \frac{\partial \log p_{\gamma}}{\partial \gamma}\\
-\frac{\partial J_{fidelity}}{\partial \beta} &\approx \frac{1}{N} \sum_{i=1}^N R(x^i,x^*) \frac{\partial \log p_{\beta}}{\partial \beta}
\end{split}
\end{align}
Since the augmenters and sequence model are mutually dependent, we adopt an alternate iterative training algorithm to update these terms as described in Algorithm~\ref{alg:TATA}.

\begin{algorithm}[!thb]
\begin{algorithmic}[0]
\Procedure{Pre-training}{}
\State Initialize model parameters $\theta, \gamma, \beta$
\State Initialize learning rate $\eta$
\State Pre-train the sequence model $\beta$ with Maximum Likelihood Estimation
\EndProcedure
\Procedure{Distribution Matching}{}
\While{Not Converged}
\State Draw random samples $(x^*, y^*) \sim D$
\If{update augmenter}
\State \# Augmenter gradient descent
\State $\gamma = \gamma - \eta \frac{\partial J_{match} + J_{fidelity}}{\partial \gamma}$ 
\State $\theta = \theta - \eta \frac{\partial J_{match} + J_{fidelity}}{\partial \theta}$
\ElsIf{update sequence-model}
\State \# Sequence model gradient descent
\State $\beta = \beta - \eta \frac{\partial J_{match} + J_{fidelity}}{\partial \beta}$
\EndIf
\State Decay learning rate $\eta$
\EndWhile
\EndProcedure
\end{algorithmic}
\caption{Distribution matching framework for Sequence-to-Sequence}
\label{alg:TATA}
\end{algorithm}
\section{Experiments}
\begin{figure*}[!t]
\begin{center}
\includegraphics[width=0.96\linewidth]{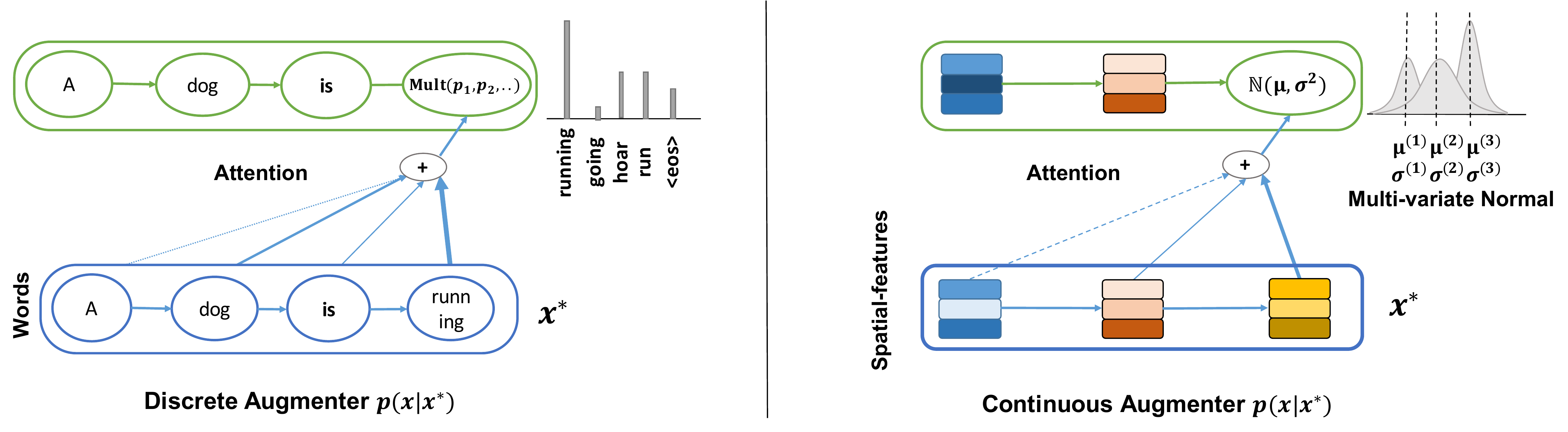}
\end{center}
   \caption{Two realizations of augmenter: the discrete augmenter uses RNN architecture to generate discrete tokens. The continuous augmenter uses the re-parametrization trick to generate continuous feature vector with a multivariate Gaussian distribution. $Mult$ denotes the multinomial distribution with categorical probability $(p_1, p_2, \cdots)$, while $\mu^{(i)}$ and $\sigma^{(i)}$ denote the mean and deviate for $i_{th}$ dimension for multi-variate Gaussian distribution.}
\label{fig:Augmenter}
\end{figure*}

In order to evaluate our distribution matching frameworks on different sequence-to-sequence applications, we select the most popular machine translation and image captioning as our benchmark. We compare our method against state-of-the-art approaches as well as MLE, RAML and RL methods. Here we design two types of augmenters as described in~\autoref{fig:Augmenter} to handle two different scenarios for machine translation and visual captioning. Our method is abbreviated as S2S-DM in the following sections. For comparability, we follow the existing papers~\cite{bahdanau2014neural,xu2015show} to adopt same network architecture, and we also apply learning rate annealing strategy described in~\cite{wu2016google} to further boost our system performance. We trained all our models on Titan X GPU, the experiments for both machine translation and visual captioning take within 3 days (excluding pre-training) to achieve the reported score. For machine translation, the performance is reported with the standard measure BLEU-4, while for image captioning, the performance is reported with CIDEr, METEOR and BLEU4 to measure different aspects of the generated captions. 
\subsection{Baseline systems}
In both experiments, we specifically compare with the following three baselines:
\begin{itemize}
    \item MLE: The maximum likelihood estimation is the de facto algorithm to train sequence-to-sequence model, here we follow~\cite{bahdanau2014neural} to train attention-based sequence-to-sequence model.
    \item RL: REINFORCE~\cite{williams1992simple} has been frequently used in sequence training to maximize the task-level metrics like~\cite{ranzato2015sequence,bahdanau2016actor}, etc. Here we design use delta BLEU as the reward function and use policy gradient to update the sequence model.
    \item RAML: We follow ~\cite{norouzi2016reward} to select the best temperature $\tau = 0.8$ in all experiments. In order to sample from the intractable payoff distribution, we adopt a stratified sampling technique described in~\cite{norouzi2016reward}. Given a ground truth $y^*$, we first sample an edit distance $m$, and then randomly select $m$ positions to replace the original labels. For each sentence, we randomly sample four candidates to perform RAML training.
\end{itemize}

\subsection{Task 1: Machine Translation}
In the machine translation experiments, we follow ~\cite{bahdanau2014neural,chung2014empirical} to design our seq-to-seq translation model. The two augmenters are also implemented with the same architecture, but they take the groundtruth tokens as their inputs. The goal of augmenters is to approximate the latent distribution with a Multinomial distribution (depicted in~\autoref{fig:Augmenter}):
\begin{gather*}
    p(x|x^*, x_{1:t-1}) = Multinomial(p)\\
    p = softmax(\lambda_t)\\
    where \qquad \lambda_t = MLP(c_t, s_t; \theta) \qquad c_t = h(x^*, s_t; \theta)
\end{gather*}
where $s_t=g(s_{t-1}, x^*, x_{t-1}; \theta)$ is the recurrent state obtained by transition function $g$ in each step and $\lambda_t$ as the softmax parameters, $p \in \mathbb{R}^V$ is the distribution over the whole vocabulary, where $c_t$ is an summarization vector from the ground truth sequence $x^*$. Here we write the derivatives as follows:
\begin{gather*}
    \frac{\partial \log p_{\theta}(x|x^*, x_{1:t-1})}{\partial \theta} = (\mathbbm{1}(x=i) - p)\frac{\partial \lambda_t}{\partial \theta}
\end{gather*}
We use $\theta$ to denote the whole parameter sets in $MLP$ layer and RNN transition function $g$.
\paragraph{IWSLT2014 German-English Dataset}
This corpus contains 153K sentences while the validation dataset contains 6,969 sentences pairs. The test set comprises dev2010, dev2012, tst2010, tst2011 and tst2012, and the total amount is 6,750 sentences. We adopt 512 as the length of RNN hidden stats and 256 as embedding size. We use the bidirectional encoder and initialize both its own decoder states and coach's hidden state with the learner's last hidden state. We pre-trained the model using MLE using a batch size of 128, we stop the pre-training stage when the dev set score converges. Then we start pre-training the augmenter using self-reconstruction, which maximizes the objective function $p(y^*|y^*)$, such pre-training procedure makes sure that the augmenter has a high initial fidelity. Finally, we train the three models jointly with distribution matching loss function. We use BLEU4~\cite{papineni2002bleu} as the evaluation metrics throughout the experiments.

\paragraph{Experimental Results}
The experimental results for IWSLT2014 German-English and English-German Translation Task are summarized in~\autoref{tab:results-IWSLT-de2en}, where we compare with MLE, RL, RAML, and many other popular competing algorithms like the reinforcement-based (MIXER and A-C), augmentation-based method (Softmax-Q) and state-of-the-art method (Transformer). From the table, we can observe that our distribution matching method outperforms all these methods and brings significant gains in both translation directions.

\begin{table}[htb]
\centering
\begin{tabular}{lll}
\toprule
\textbf{Model} & \textbf{DE2EN} & \textbf{EN2DE}\\
\midrule
MIXER~\cite{ranzato2015sequence} & 21.81 &  - \\
BSO~\cite{wiseman2016sequence} & 26.36 & -\\
A-C~\cite{bahdanau2016actor} & 28.53  & -\\
Softmax-Q~\cite{ma2017softmax} & 28.77 & -\\
Transformer~\cite{vaswani2017attention} & 30.21 & 25.02 \\
\midrule
MLE & 29.10 &  24.40 \\
RL & 29.70 & 24.75 \\
RAML &  29.47 & 24.86 \\
S2S-DM & \textbf{30.92} & \textbf{25.54} \\
\bottomrule
\end{tabular}
\caption{Experimental results on IWSLT-2014 German-English Machine Translation Task}
\label{tab:results-IWSLT-de2en}
\end{table}

\subsection{Task 2: Image Captioning}
\begin{figure}[htb]
\begin{center}
\includegraphics[width=0.95\linewidth]{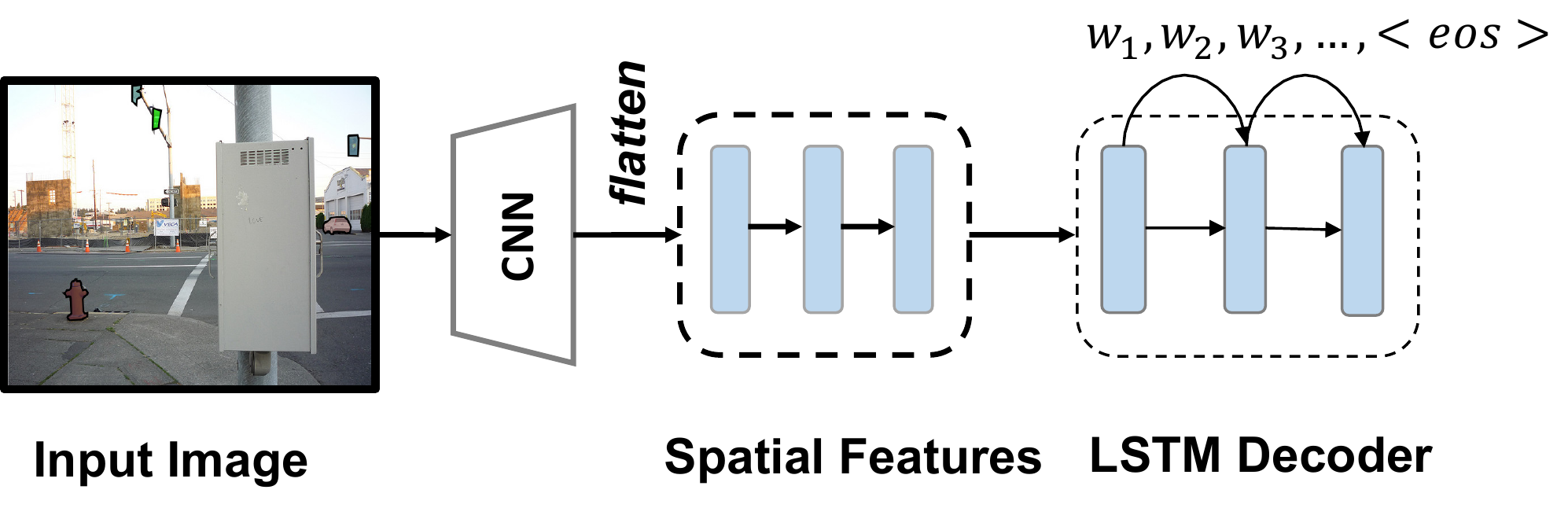}
\end{center}
   \caption{The sequence-to-sequence architecture flattens the image CNN feature map as a sequence to fully represent the image's spatial features.}
\label{fig:image-captioning}
\end{figure}

In the image captioning experiments, we follow~\cite{DBLP:conf/cvpr/RennieMMRG17,xu2015show} to design our seq-to-seq captioning model as depicted in~\autoref{fig:image-captioning}. The two augmenters are also based on similar architecture (depicted in~\autoref{fig:Augmenter}), but the source augmenter takes the visual representation as input while the target augmenter takes the groundtruth tokens as inputs (same as MT augmenter). For source augmenter, we use the re-parameterization trick to denote the continuous visual representation as a multi-variate Gaussian distribution $p(x_t|x^*, x_{1:t-1}) \sim \mathcal{N}(x^*_t, \sigma_t^2)$. By assuming independence between dimensions, we can simplify the standard deviate $\sigma_t$ as $diag(\lambda^1_t, \cdots, \lambda^F_t)$. Hence, the output probability distribution can be written as follows:
\begin{gather*}
\log p_{\theta}(x|x^*) = \sum_{t=1}^{t=T} \sum_{k=1}^{k=K} [-\frac{1}{2} \log(2 \pi {\lambda_t^k}^2) - \frac{(x_t^k - {x_t^*}^k)^2}{2  {\lambda_t^k}^2}]
\end{gather*}
We here adopt re-parameterization trick $x_t = x^*_t + \tilde{n}_{t}$ and use an RNN to predict the deviate at each time step with $\tilde{n}_{t} = \lambda_t \odot n_t$ and $\lambda_t = MLP(s_t;\theta)$. The noise $n_t \sim \mathcal{N}(0, \mathbb{I})$ is sampled from isotropic Gaussian and the RNN hidden state  $s_t$ is obtained from transition function $s_t = g(c_{t-1}, s_{t-1}, \tilde{n}_{t-1};\theta)$. We use $k$ to denote the individual dimension and formally write its derivatives as follows:
\begin{align*}
\frac{\partial \log p_{\theta}(x|x^*)}{\partial \theta} = \sum_{t=1}^{t=T} \sum_{k=1}^{k=K} \frac{1}{\lambda_t^k} \frac{\partial \lambda_t^k}{\partial \theta} = \sum_{t=1}^{t=T} \sum_{k=1}^{k=K} \frac{\partial \log \lambda_t^k}{\partial \theta}
\end{align*}
where $K$ represents the multi-variate Gaussian dimension, and $T$ represents the length of the sequence. We use $\theta$ to denote the whole parameter sets in both $MLP$ and RNN $g$.
\paragraph{MSCOCO Dataset}
We evaluate the performance of our model on MS-COCO captioning dataset~\cite{lin2014microsoft}. The MS-COCO dataset contains 123,287 images for training and validation, 40775 images for testing. Here we use the standard split described by Karpathy\footnote{\small \url{https://github.com/karpathy/neuraltalk2}} for which 5000 images were used for both validation and testing and the rest for training. We pre-train the model on this data using a batch size of 256 and validate on an out-of-domain held-out set, this stage is ended when the validation score converges or the maximum number of epochs is reached. After pre-training, we continue distribution matching training on the original paired dataset. The LSTM hidden, image, word and attention embeddings dimension are fixed to 512 for all of the models discussed herein. We initialize all models by training the model under the cross-entropy objective with a learning rate of $5 \times 10^{−4}$. We anneal the learning rate by a factor of 0.8 every three epochs. At test time, we do beam search with a beam size of 4 to decode words until the end sentence symbol is reached. We use different standard evaluation metrics described in~\cite{chen2015microsoft}, including BLEU@N~\cite{papineni2002bleu}, METEOR, and CIDEr to measure different aspects of generated captions. 

\begin{table}[htb]
\small
\centering
\begin{tabular}{lccc}
\toprule
\textbf{Model} & \textbf{CIDEr} & \textbf{BLEU} & \textbf{MET} \\
\midrule
\makecell[l]{Neuraltalk2} & 66.0 & 23.0 & 19.5 \\
\midrule
\makecell[l]{Soft-Attention~\cite{xu2015show}} & 66.7 & 24.3 & 23.9 \\
\midrule
\makecell[l]{Att2in SCST~\cite{DBLP:conf/cvpr/RennieMMRG17}} & 111.4 & 33.3 & 26.3 \\
\midrule
\makecell[l]{Att2in MLE~\cite{DBLP:conf/cvpr/RennieMMRG17}} & 101.3 & 31.3 & 26.0 \\
\midrule
\makecell[l]{Att2in RL~\cite{DBLP:conf/cvpr/RennieMMRG17}} & 109.8 & 32.8 & 26.0 \\
\midrule
Att2in RAML & 98.5 & 31.2 & 26.0 \\
\midrule
S2S-DM & \textbf{112.8} & \textbf{33.9} &  \textbf{26.4}\\
\bottomrule
\end{tabular}
\caption{Experimental results on image captioning task, where we use CIDEr, BLEU4 and METEOR (MET) to measure the system performance.}
\label{tab:captioning}
\end{table}

\begin{table*}[!thb]
\centering
\begin{tabular}{lll}
\toprule
\multirow{2}{*}{\textit{\begin{tabular}[c]{@{}c@{}}Synonym \\ Replacement\end{tabular}}} & Reference                                                  & taihsi natives seeking work ... being \textbf{hired}, and later their colleagues maintain ... \\
                                                                                         & Sample & taihsi natives seeking work ... being \textbf{employed}, and later their colleagues maintain ... \\ \midrule
\multirow{2}{*}{\textit{Simplification}}                                                 & Reference                                                  &  i once took mr tung ... that a narrow alley \textbf{could have accommodated so many people}. \\
                                                                                         & Sample & i once took mr tung ... that a narrow alley \textbf{have a lot of people}. \\ \midrule
\multirow{2}{*}{\textit{Re-Ordering}}                                                    & Reference                                                  & \textbf{he and I} went to the theater yesterday to see a film. \\
                                                                                         & Sample & \textbf{I and he} went to the theater yesterday to see a film. \\ \midrule
\multirow{2}{*}{\textit{Repetition/Missing}}                                                          & Reference                                                  & \textbf{and} i had recently discovered a \textbf{bomb shelter} ... \\
                                                                                         & Sample & i have discovered a \textbf{place place} ... \\ 
\bottomrule
\end{tabular}
\caption{Samples drawn from Augmenter in LDC Chinese-English Translation task}
\label{tab:cluster-example}
\end{table*}

\paragraph{Experimental Results}
We summarize the experimental results in~\autoref{tab:captioning}, where we mainly compare with MLE, RL, and RAML. We implement our Att2in RAML and S2S-DM based on the open repository\footnote{\small \url{https://github.com/ruotianluo/self-critical.pytorch}}. As can be seen, our method achieves remarkable gains across different metrics over RL, MLE and RAML, besides, our single model best results also slightly outperform SCST training algorithm~\cite{DBLP:conf/cvpr/RennieMMRG17}. These results have consistently demonstrated the advantage of distribution matching algorithm under continuous sequence scenarios, which can be potentially extended to more vision-related sequence-to-sequence tasks. 

\subsection{Results Analysis}
\begin{figure}[htb]
\begin{center}
\includegraphics[width=1.0\linewidth]{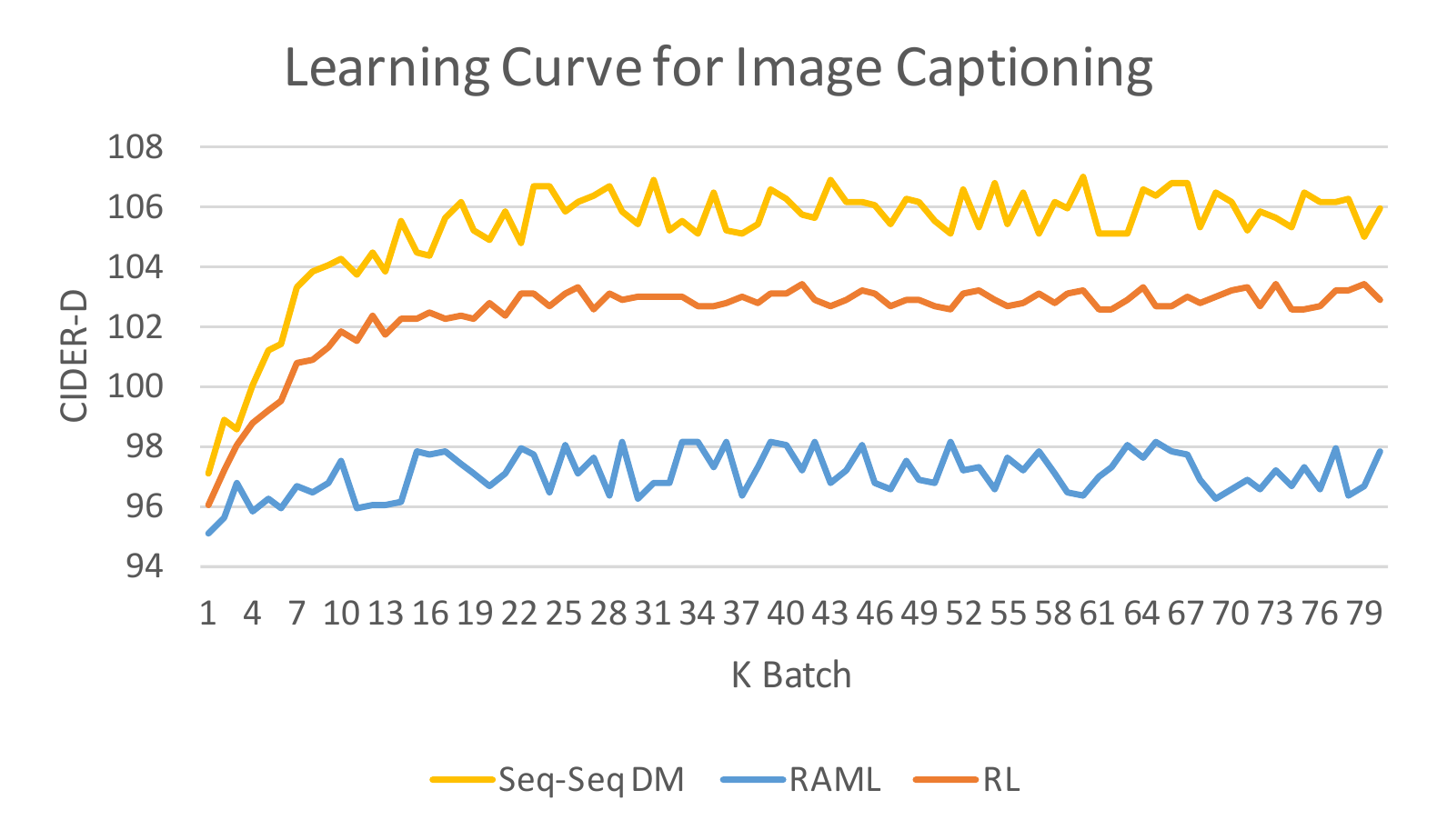}
\end{center}
   \caption{Learning curves for machine translation.}
\label{fig:lc_IC}
\end{figure}

\begin{figure}[htb]
\begin{center}
\includegraphics[width=1.0\linewidth]{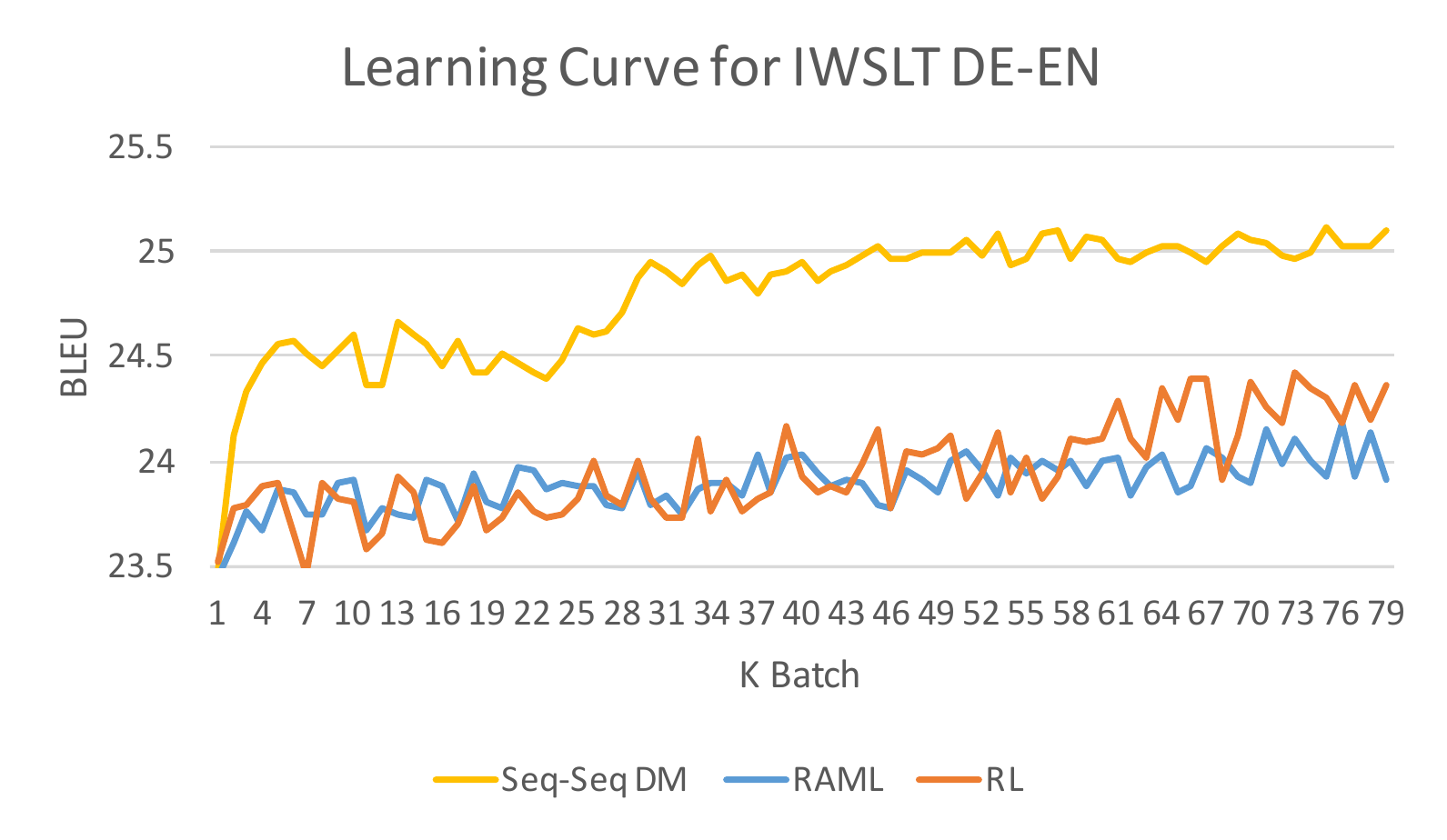}
\end{center}
   \caption{Learning curves for machine translation.}
\label{fig:lc_MT}
\end{figure}

\begin{figure}[htb]
\begin{center}
\includegraphics[width=1.0\linewidth]{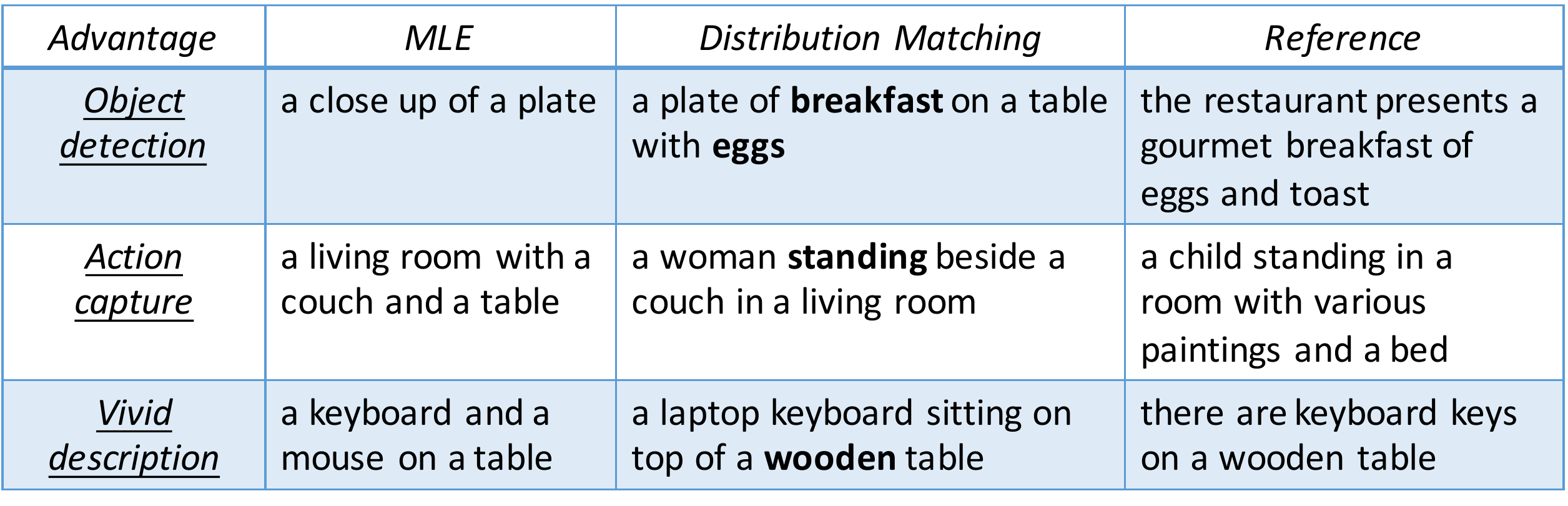}
\end{center}
   \caption{Examples of captions on MS-COCO dataset, we compare with the samples generated with MLE method.}
\label{fig:Caption-example}
\end{figure}

From the above results, we can observe limited improvements yielded by the RAML algorithm on most tasks and even causes performance degradation in some tasks (LDC Chinese-English, Image Captioning). We conjecture that it's caused by the heuristic strategic replacement strategy which breaks both the semantic and structure information. Especially in image captioning, there already exist five references for the target side, further augmenting the target site receives very little gain. For reinforcement learning, it only focuses on enhancing the target-side decision process, while our method to augment the source sequence is able to expose the model to more unseen source-side sequences. Such advantage makes our model better in handling unseen visual representation and generalizing test cases. We empirically verify the effectiveness of S2S-DM algorithm on augmenting both the discrete and continuous sequence data pairs.
\paragraph{Learning Curves}
Here we showcase the learning curves of the sequence-to-sequence model for both the IWSLT machine translation task and image captioning task separately in~\autoref{fig:lc_IC} and~\autoref{fig:lc_MT}. We can observe very stable improvements of our distribution matching algorithm over the pre-trained model. In machine translation task, RL and RAML can both boost the model by 0.5-0.8 BLEU, while distribution matching can boost roughly 1.5 BLEU. In image captioning, RAML does not benefit the training evidently, while RL and distribution matching both improve the performance remarkably in terms of CIDEr-D.\\
\paragraph{Case Studies}
In order to give a more intuitive view of latent distribution approximated by our augmenters, we here draw some high-probability samples from the augmenters. We can observe that most of the sample pairs remain their fidelity to the original pair, their modifications against the original ground truth are mainly classified into four types, which we demonstrate in~\autoref{tab:cluster-example}. Though the augmenter introduces some noises into the references, these noises are still under control, and the most frequent noises are missing and repetition. Further, we also demonstrate a few image captioning examples in~\autoref{fig:Caption-example} to showcase the advantage of our distribution-matching framework. As can be seen, the generated samples adopt a more vivid and diverse language expression. More detailed descriptions about the objects in the picture are included.

\section{Conclusion}
In this paper, we propose a new end-to-end training algorithm to resolve the data sparsity problem in sequence-to-sequence applications. We have verified the capability of our model in two popular applications (machine translation and image captioning) to understand more diverse inputs and generate more complicated outputs. We look forward to testing our algorithms on more sequence-to-sequence applications to verify its generality.

\bibliography{aaai}
\bibliographystyle{aaai}

\appendix
\onecolumn
\section{Supplementary Material}
\subsection{LDC Chinese-English Dataset}
The LDC Chinese-English training corpus consists of 1.25M parallel sentence, 27.9M Chinese words and 34.5M English words. We choose NIST 2003 as our development set and evaluate our results on NIST 2005, NIST2006. We adopt a similar setting as IWSLT German-English translation task, we use 512 as hidden size for GRU cell and 256 as embedding size. The experimental results for LDC Chinese-English translation task are listed in~\autoref{tab:results-NIST-de2en}.
\begin{table}[htb]
\centering
\begin{tabular}{lll}
\toprule
\textbf{Model} & \makecell{\textbf{CH2EN} \\NIST03/05/06} & \makecell{\textbf{EN2CH}\\NIST03/05/06} \\
\midrule
MLE & 39.0 / 37.1 / 39.1 & 17.57 / 16.38 / 17.31\\
RL & 41.0 / 39.2 / 39.3 & 18.44 / 16.98 / 17.80\\
RAML &  40.2 / 37.3 / 37.2 &  17.83 / 16.52 / 16.79 \\
S2S-DM  & \textbf{41.8 / 39.3 / 39.5} & \textbf{18.92 / 17.36 / 17.88} \\
\bottomrule
\end{tabular}
\caption{Experimental results on NIST Chinese-English Machine Translation Task}
\label{tab:results-NIST-de2en}
\end{table}

\subsection{Augmenter Results Visualization}
\begin{figure*}[!htb]
\begin{center}
\includegraphics[width=0.90\linewidth]{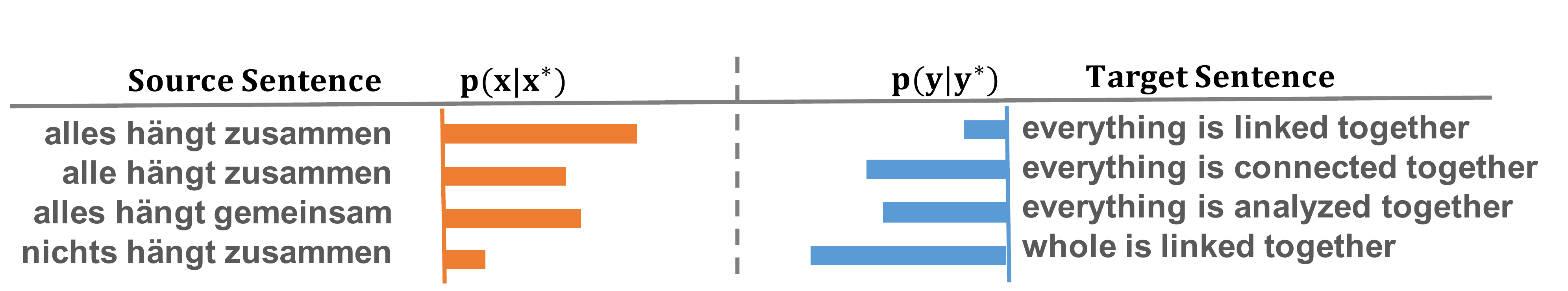}
\end{center}
   \caption{Paired samples drawn from augmenters in IWSLT DE-EN translation task.}
\label{fig:paired samples}
\end{figure*}

Such observation has confirmed our intuition to build a semantic/syntactic preserving cluster around the ground truth. We here showcase the paired augmentation samples in~\autoref{fig:paired samples}. 
\end{document}